\documentclass{llncs}
\usepackage[T1]{fontenc}
\usepackage{graphicx}
\usepackage{amsmath,amssymb,amsfonts}
\usepackage{subfigure}
\usepackage{hyperref}
\hypersetup{hidelinks}

\begin{document}
\title{DensePed-Lite: Quality-Aware Adaptive Detection for Dense Pedestrians under Occlusion}
\titlerunning{DensePed-Lite}
\author{ZiAn Wang\inst{1} \and
MingZhe Liu\inst{2} \and
Chaoyi Guo\inst{3} \and
ChangChun Li\inst{1,4} \and
Fangming Gu\inst{1,4}\thanks{Corresponding author.}}
\authorrunning{ZiAn Wang et al.}
\institute{Jilin University, China\\
\email{wangza2124@mails.jlu.edu.cn}
\and
Shenzhen University, China\\
\email{cnmingzheliu@outlook.com}
\and
Taiyuan University of Technology, China\\
\email{13293974463@163.com}
\and
Key Laboratory of Symbolic Computation and Knowledge Engineering of Ministry of Education, Jilin University, China\\
\email{changchunli93@gmail.com, gufm@jlu.edu.cn}}
\maketitle              
\begin{abstract}
Pedestrian detection plays a crucial role in computer vision with applications in autonomous driving, surveillance, and public safety. However, real-world dense scenes bring severe challenges, including heavy occlusion, drastic scale variations, and strict real-time requirements. Existing lightweight detectors struggle to balance accuracy and efficiency while often neglecting quality-aware feature modeling and consistency between classification and localization, leading to unstable performance under crowded conditions. To address these issues, we propose DensePed-Lite, a unified framework built on a single principle: under occlusion the network should adapt its behavior to the quality of what it observes rather than assume complete information. This principle is realized at three points where occlusion does the most damage---unreliable confidence scoring (UQE), fragmented spatial coverage (MPSC), and incoherent multi-scale fusion (CTDM)---so that the three mechanisms reinforce one another instead of acting in isolation, all without significantly increasing complexity. Experiments on CityPersons and CrowdHuman validate that DensePed-Lite achieves superior accuracy-efficiency trade-off compared with recent state-of-the-art lightweight methods, making it suitable for real-time deployment in dense pedestrian scenarios.

\end{abstract}

\section{Introduction}
\label{sec:intro}

Pedestrian detection is fundamental in computer vision with applications in autonomous driving, surveillance, and public safety \cite{dalal2005hog}. Crowded scenes pose challenges: severe occlusion, large scale variations, and real-time requirements significantly impact accuracy and limit deployment.

Research has progressed through three directions: \textbf{occlusion handling} via refined suppression and fusion \cite{liu2019adaptive,fang2024improved}; \textbf{multi-scale modeling} through hierarchical features \cite{tan2020efficientdet,liu2024yolov8cb}; \textbf{lightweight architectures} balancing accuracy and efficiency \cite{sun2025dfine,liu2025deim}. However, enhanced interaction increases overhead, parameter reduction compromises occluded-pedestrian detection, and attention limits real-time deployment.

These trade-offs reflect a deeper issue: existing lightweight detectors optimize for complete information but lack mechanisms to preserve task-relevant signals when occlusion introduces ambiguity. This manifests in three failures: unreliable confidence scores suppress partially visible pedestrians, fragmented spatial features corrupt multi-scale representations, and limited scanning paths cannot recover from occlusion-induced information loss. Current methods address symptoms separately without recognizing all three stem from the same issue: information loss under ambiguity.

We build DensePed-Lite on one principle: under occlusion the network should adapt to the quality of what it observes rather than assume complete input. We apply this at three failure points. At the detection head, UQE makes scoring sensitive to localization uncertainty so confident but poorly localized boxes are not trusted blindly. In the backbone, MPSC restores spatial coverage through complementary scanning paths gated by reliability. Before fusion, CTDM stabilizes and verifies features across scales so conflicting signals are reconciled rather than averaged. All three act on observation quality, reinforcing one another: cleaner spatial coverage yields meaningful uncertainty estimates, filtered predictions support coherent fusion, and coherent features guide path selection.

The main contributions are:
\begin{itemize}
\item We identify that lightweight detector failures in dense scenes stem from inability to adapt behavior under occlusion-induced ambiguity, manifesting as three coupled failure modes (confidence, spatial coverage, scale coherence) that reinforce one another. We design a unified quality-aware adaptive framework that breaks this cycle at all three points simultaneously.

\item We propose three novel mechanisms realized at critical failure points: UQE explicitly models localization uncertainty through distribution variance with learnable penalty weighting; MPSC provides bidirectional spatial redundancy with adaptive quality-aware gating; CTDM enforces multi-scale coherence through a stabilize-verify-refine closed loop. These mechanisms are designed to reinforce one another rather than act independently.

\item Experiments on CityPersons and CrowdHuman show a strong accuracy-efficiency trade-off against recent lightweight methods, while comprehensive ablations validate the coupled effects of the three mechanisms.
\end{itemize}

\section{Related Work}
\label{sec:related}

\subsection{Occlusion Handling}
Occlusion conceals critical visual cues in dense scenes. Adaptive NMS \cite{liu2019adaptive} refines suppression for overlapping detections; attention-based methods \cite{fang2024improved} focus on discriminative regions; multi-modal fusion \cite{zhou2025fusion} improves robustness through complementary modalities. These methods often rely on local features or lack semantic context, limiting robustness under heavy occlusion. We explicitly model prediction uncertainty for automatic down-weighting of unreliable detections.

\subsection{Multi-scale Modeling}
Robust multi-scale representation is essential for large scale variations. BiFPN \cite{tan2020efficientdet} introduces bidirectional pyramids with learnable weights; YOLOv8-CB \cite{liu2024yolov8cb} enhances attention at multiple scales; joint modeling \cite{wang2025datafusion} incorporates auxiliary tasks like crowd counting. Despite advances, feature misalignment and boundary ambiguity limit accuracy. Our CTDM enforces coherence by stabilizing distributions and verifying cross-scale consistency before fusion.

\subsection{Lightweight Detection Methods}
The YOLO series has evolved with YOLOv5--YOLOv13 \cite{khanam2024yolov5,varghese2024yolov8,wang2024yolov10,yolo11,zhang2025yolov12,wang2025yolov13} introducing progressive improvements. Specialized methods \cite{li2024gryolo,liu2025mscd,li2024largekernel,ding2023swyolox} enhance features but increase cost. DETR-based approaches \cite{sun2025dfine,liu2025deim} show promise through refined modeling, though requiring more parameters. DensePed-Lite addresses trade-offs through quality-aware mechanisms that preserve task-relevant signals under occlusion.

\section{Method}

\subsection{Overall Framework}

Existing lightweight detectors optimize for complete visibility but lack mechanisms to preserve task-relevant signals when occlusion introduces ambiguity. This manifests in three interconnected failures: classification remains overconfident while localization degrades, causing NMS to suppress partially visible pedestrians; fragmented spatial features corrupt multi-scale representations as fusion aggregates contradictory signals without verification; single-path scanning with fixed receptive fields cannot recover from occlusion-induced information loss. These failures share a common root---\emph{information loss under ambiguity}---and form a vicious cycle.

Our key insight is that occlusion-robustness requires \emph{quality-aware adaptive processing} where the network dynamically adjusts based on input quality. DensePed-Lite instantiates this through three mechanisms breaking the failure cycle: UQE models prediction uncertainty via variance, enabling NMS to down-weight unreliable detections; MPSC provides spatial redundancy through multi-path scanning with adaptive gating; CTDM enforces multi-scale coherence by stabilizing feature distributions and verifying consistency before fusion. These create a virtuous cycle where complete spatial coverage enables meaningful uncertainty estimation, filtered predictions enable coherent fusion, and coherent features guide path selection.

\begin{figure}[t]
\centering\includegraphics[width=\textwidth]{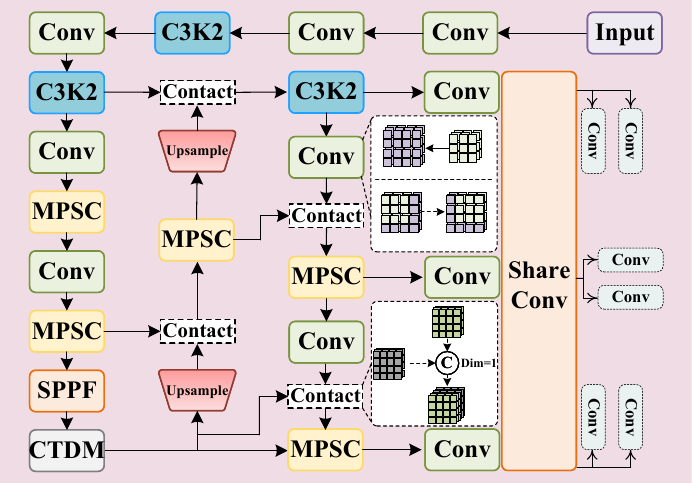}
\caption{The architecture of the network.}
\label{fig:network}
\end{figure}

We implement our approach based on YOLOv11n \cite{yolo11}, with MPSC at P4/P5, CTDM integrated with SPPF, and UQE replacing the standard detection head. Figure~\ref{fig:network} illustrates this architecture.

\subsection{UQE: Uncertainty-Aware Quality Estimation}

Standard NMS relies on classification confidence, but this fails under occlusion. Consider a heavily occluded pedestrian where only the head is visible: the classification branch confidently predicts ``person'' (high $S_{cls}$) while the regression branch produces uncertain box coordinates. NMS treats this high-confidence, low-quality detection as reliable, causing false suppression. A natural remedy is to read a quality cue from the regression distribution itself, for instance by pooling its TopK responses \cite{li2023generalizedfocal}. However, TopK pooling captures only how sharp the distribution is and ignores how widely it spreads, so an occluded box with a sharp peak but a wide spread receives a score similar to a clean one. What NMS actually needs is a signal that grows with localization uncertainty.

We therefore model uncertainty explicitly rather than reading it off implicit features, directly incorporating prediction variance into the score:
\begin{equation}
S_{final} = \sigma\left(S_{cls} + f_\theta(\text{TopK}(B_{reg})) - \lambda \cdot \text{Var}(B_{reg})\right)
\end{equation}
where $\text{Var}(B_{reg})$ quantifies localization uncertainty, $\lambda$ is learnable, and subtraction enforces monotonic penalty. Three design choices make this an uncertainty model rather than another feature: subtraction rather than concatenation encodes the prior that uncertainty should reduce confidence; a learnable $\lambda$ adapts the penalty to the occlusion characteristics of each dataset; and pairing it with depthwise separable convolutions keeps the head efficient.

To keep the shared head lightweight, its convolutions are factorized into depthwise and pointwise stages \cite{howard2017mobilenets}:
\begin{equation}
Y_m = \sum_{k=1}^{C_{in}} (X_k * D_k) \cdot P_{k,m}, \quad m = 1, 2, \ldots, C_{out}
\end{equation}
Decoupled branches separate classification and regression, with DFL producing distributions for variance computation. Variance adds only $\mathcal{O}(N \times C_{reg})$ operations ($N$ predictions, $C_{reg}=4$), enabling quality-aware NMS with minimal computational overhead.

\begin{figure}[t]
\centering\includegraphics[width=\textwidth]{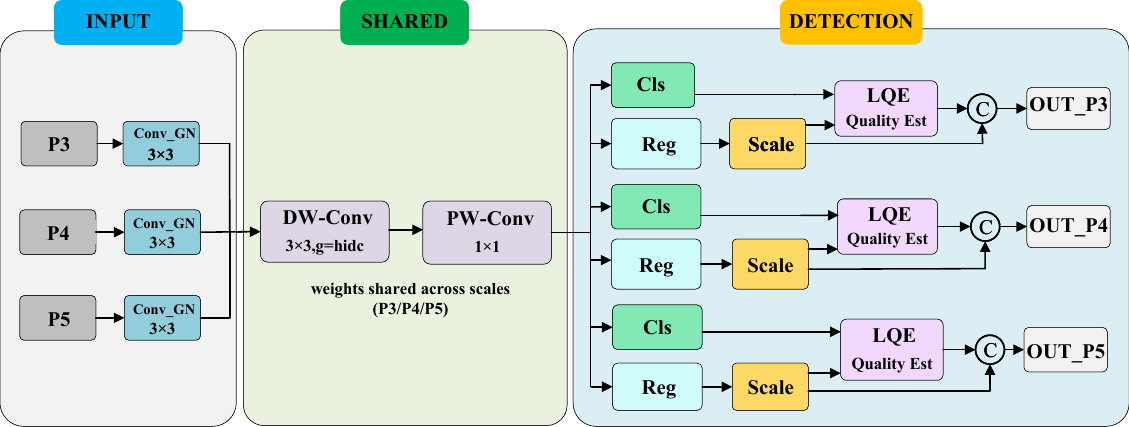}
\caption{The architecture of the UQE module.}
\label{fig:uqe}
\end{figure}

\subsection{MPSC: Multi-Path Spatial Completion}

Lightweight backbones employ sequential convolutions with fixed receptive fields. When occlusion blocks the scanning path, entire regions become unobservable with no recovery mechanism. Larger kernels retain single paths; multi-scale fusion provides scale diversity but not spatial path diversity; self-attention costs prohibit lightweight deployment. What is needed is spatial redundancy through complementary scanning directions.

We restore this redundancy without abandoning the lightweight backbone by equipping the C3k2 multi-branch framework with a pair of complementary scanning paths and an adaptive gating scheme that decides how much to trust each path. Three design choices make this effective. First, bidirectional spatial scanning with standard and shifted patterns provides directional coverage:
\begin{equation}
X_{\text{scan-std}} = \mathrm{Scan}(X, I_{std}), \quad
X_{\text{scan-shift}} = \mathrm{Scan}(X, I_{shift})
\end{equation}
When a pedestrian is occluded from the left, the standard scan misses the left boundary but the shifted scan observes the right boundary and infers context. Second, learnable gating $\alpha, \beta$ enables quality-aware path selection:
\begin{equation}
Y_1 = X \odot \alpha + \mathrm{DropPath}(\mathrm{VMM}(\mathrm{LN}(X)))
\label{eq:res1}
\end{equation}
\begin{equation}
Y_2 = Y_1 \odot \beta + \mathrm{MLP}(\mathrm{LN}(Y_1))
\label{eq:res2}
\end{equation}
The network learns $\alpha \to 1$ for high-quality inputs and $\alpha \to 0$ for heavy occlusion, forming implicit quality-awareness that complements UQE's explicit uncertainty quantification. Third, strategic P4/P5 deployment reduces overhead on smaller feature maps (20$\times$20 vs. 80$\times$80) while enhancing effectiveness at semantic levels.

Weighted fusion aggregates multiple paths:
\begin{equation}
F_{\text{final}} = \sum_{i=1}^{n} W_i \cdot \mathrm{Path}_i(F_{i-1})
\label{eq:fusion}
\end{equation}
where $W_i$ adapts based on path completeness, down-weighting occluded paths while emphasizing complete observations.

\subsection{CTDM: Multi-Scale Coherence Preservation}

Feature pyramids assume different scales provide complementary views, but occlusion breaks this: P3 detects partial edges, P4 sees fragmented shapes, P5 observes only background. Standard fusion aggregates these contradictory signals without verification, producing unstable predictions. The root cause is assuming rather than enforcing coherence.

We enforce coherence through a three-stage pipeline that stabilizes, verifies, then refines. We first stabilize the skewed feature distributions that occlusion induces through dynamic activation normalization, preventing drifting activations from corrupting subsequent processing. We then verify global consistency by modeling dependencies over feature statistics rather than raw activations, which naturally suppresses contributions from unreliable occluded regions:
\begin{equation}
G = X_{\text{enh}} \oplus
\mathrm{Reshape}\Big(\mathrm{TSSA}\big(\mathrm{F\&T}(\mathrm{DyT}_1(X_{\text{enh}}))\big)\Big)
\label{eq:ctdm1}
\end{equation}
Finally, we refine local details under guidance of the verified global context through multi-scale local attention, sharpening boundaries where they remain ambiguous:
\begin{equation}
E = \mathrm{Mona}_2\Big( G \oplus
\mathrm{FFN}\big(\mathrm{DyT}_2(\mathrm{Mona}_1(G))\big)\Big) \oplus G
\label{eq:ctdm2}
\end{equation}
The residual connections close the loop: the verified global context anchors local refinement, and the refined features flow back to stabilization, so that confident signals from clear scales progressively correct the ambiguous ones rather than being averaged against them.

A final dual-path fusion preserves the original local details alongside the refined representation:
\begin{equation}
Y = \mathrm{Conv}_{1\times1}\Big(\mathrm{Concat}[X_{\text{loc}},\,E]\Big)
\label{eq:ctdm3}
\end{equation}
This stabilize-verify-refine loop lets the three stages reinforce one another: stable distributions make global modeling reliable, reliable global context guides local refinement, and the refined details in turn yield cleaner distributions for the next layer, all while keeping the attention lightweight enough for real-time use.

\begin{figure}[t]
\centering\includegraphics[width=\textwidth]{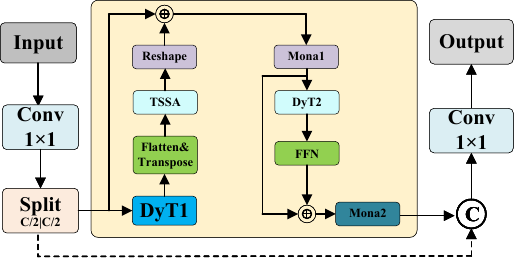}
\caption{The architecture of the CTDM module.}
\label{fig:ctdm}
\end{figure}

\subsection{Model Complexity Analysis}

The UQE module achieves significant complexity reduction through depthwise separable convolutions, decreasing from $\mathcal{O}(C_{in} \times C_{out} \times K^2 \times H \times W)$ to $\mathcal{O}(C_{in} \times K^2 \times H \times W + C_{in} \times C_{out} \times H \times W)$, yielding a reduction factor of $\frac{K^2}{C_{out} + K^2}$. This decomposition reduces parameters from $\Theta(C_{in} \times C_{out} \times K^2)$ to $\Theta(C_{in} \times K^2 + C_{in} \times C_{out})$, achieving substantial savings when $K^2 < C_{out}$. The MPSC and CTDM modules introduce minimal overhead with $\mathcal{O}(L \times D^2)$ and $\mathcal{O}(N \times C^2)$ complexity respectively, controlled through strategic deployment at lower-resolution feature maps. These improvements enable DensePed-Lite to maintain competitive accuracy while significantly reducing computational requirements for real-time dense pedestrian detection.

\begin{table}[t]
\caption{Comparison Results on CrowdHuman and CityPersons Datasets. Boldface indicates the best result and underlining indicates the second-best result among ultra-lightweight methods (Params $\leq$ 4M), respectively.}
\label{tab:crowd_city}
\centering
\resizebox{\textwidth}{!}{%
\begin{tabular}{lcccc|cccccc}
\hline
 & \multicolumn{4}{c|}{\textbf{CrowdHuman}} & \multicolumn{4}{c}{\textbf{CityPersons}} & \textbf{GFLOPs} & \textbf{Params} \\
\cline{2-9}
\textbf{Method} & \textbf{Recall} & \textbf{AP50(\%)} & \textbf{AP(\%)} & \textbf{Precision} & \textbf{Recall} & \textbf{AP50(\%)} & \textbf{AP(\%)} & \textbf{Precision} &  &  \\
\hline
\multicolumn{11}{l}{\textit{Baseline}} \\
\hline
YOLOv11n \cite{yolo11} & 0.678 & 79.2 & 48.4 & 0.838 & 0.513 & 59.6 & 36.1 & 0.760 & 6.3   & 2.58M \\
\hline
\multicolumn{11}{l}{\textit{Traditional Lightweight Detectors}} \\
\hline
YOLOv5n \cite{khanam2024yolov5}  & 0.658 & 77.6 & 47.4 & 0.833 & 0.502 & 59.7 & 36.6 & 0.789 & 7.1   & 2.5M \\
YOLOv8n \cite{varghese2024yolov8} & 0.666 & 79.0 & 48.7 & 0.842 & 0.518 & 60.8 & 37.3 & 0.783 & 8.1   & 3.0M \\
YOLOv10n \cite{wang2024yolov10}   & 0.658 & 78.3 & 48.2 & 0.829 & 0.502 & 60.4 & 36.9 & 0.776 & 6.5   & 2.3M \\
\hline
\multicolumn{11}{l}{\textit{Recent Ultra-Lightweight SOTA Methods (Params $\leq$ 4M)}} \\
\hline
YOLO26n \cite{yolo26}              & 0.671 & 77.0 & 44.3 & 0.814 & 0.496 & 59.5 & 35.9 & 0.768 & \textbf{5.4}   & 2.44M \\
YOLOv12n \cite{zhang2025yolov12}   & 0.674 & 79.4 & 48.9 & 0.847 & 0.516 & 60.3 & 37.2 & 0.769 & 6.5   & 2.52M \\
YOLOv13n \cite{wang2025yolov13}    & 0.671 & 79.2 & 48.6 & 0.845 & 0.509 & 59.6 & 36.4 & 0.763 & 6.4   & 2.45M \\
YOLOv8-CB \cite{liu2024yolov8cb}   & \underline{0.706} & \underline{80.1} & 48.5 & -- & -- & -- & -- & -- & 7.5   & 2.70M \\
Hyper-YOLO-n \cite{feng2024hyper}  & 0.703 & 81.0 & 50.4 & 0.846 & 0.503 & 59.7 & 36.5 & 0.806 & 9.7  & 3.63M \\
D-FINE-N \cite{sun2025dfine}       & 0.682 & 79.8 & 49.2 & 0.850 & 0.521 & 61.2 & 38.4 & 0.782 & 7.0   & 4.0M \\
DEIM-D-FINE-N \cite{liu2025deim}   & 0.688 & \underline{80.1} & \underline{49.8} & \underline{0.851} & \underline{0.525} & \underline{61.8} & \textbf{38.8} & \underline{0.789} & 7.0   & 4.0M \\
\hline
\multicolumn{11}{l}{\textit{\textbf{Proposed Method}}} \\
\hline
\textbf{DensePed-Lite (Ours)} & \textbf{0.713} & \textbf{80.9} & \textbf{50.3} & \textbf{0.858} & \textbf{0.529} & \textbf{62.1} & \underline{38.6} & \textbf{0.797} & \underline{5.5} & \textbf{2.53M} \\
\hline
\multicolumn{11}{l}{\textit{Higher-Capacity Lightweight Methods}} \\
\hline
Mamba-YOLO-T \cite{wang2025mambayolo} & 0.717 & 81.8 & 51.7 & 0.857 & 0.508 & 60.9 & 37.8 & 0.812 & 12.4  & 5.66M \\
\hline
\multicolumn{11}{l}{\textit{Large-scale Models}} \\
\hline
YOLOv11s \cite{yolo11}           & 0.719 & 81.4 & 52.9 & 0.864 & 0.541 & 63.1 & 39.9 & 0.811 & 21.3  & 9.4M \\
RT-DETR-R18 \cite{zhao2024rtdetr} & 0.704 & 80.8 & 51.6 & 0.851 & 0.534 & 62.8 & 38.5 & 0.803 & 56.9  & 19.9M \\
\hline
\multicolumn{11}{l}{\textit{Traditional Detectors}} \\
\hline
SSD \cite{liu2016ssd}              & 0.602 & 71.6 & 36.7 & 0.810 & 0.428 & 50.1 & 32.7 & 0.767 & 86.0  & 23.7M \\
Faster R-CNN \cite{ren2017faster}  & 0.680 & 78.6 & 49.5 & 0.846 & 0.514 & 60.2 & 36.4 & 0.773 & 251.4 & 41.3M \\
RetinaNet \cite{lin2017focal}      & 0.654 & 76.1 & 46.7 & 0.838 & 0.509 & 60.8 & 37.5 & 0.788 & 199.7 & 46.4M \\
\hline
\end{tabular}%
}
\end{table}

\section{Experiments}

\subsection{Experimental Setting}

\subsubsection{Datasets.}
We use CityPersons \cite{zhang2017citypersons} (5,000 urban street images: 2,975 train, 500 val, 1,525 test) and CrowdHuman \cite{shao2018crowdhuman} ($\sim$15,000 train, 4,370 val images with over 470,000 instances, averaging 23 pedestrians per image with severe occlusion).

\subsubsection{Metrics.}
We follow COCO protocol using AP@0.50 (AP50), AP@0.50:0.95 (AP), Recall, and Precision for accuracy; GFLOPs and Parameters for efficiency.

\subsubsection{Implementation Details.}
PyTorch on NVIDIA RTX 4090 (24GB). Training: 200 epochs, SGD (momentum 0.937, lr 0.01, weight decay 0.0005), batch size 32, 640$\times$640 input with Mosaic augmentation. Results averaged over three runs.

\subsection{Experimental Results}

\subsubsection{Overall Performance.}
We evaluate DensePed-Lite against representative detectors across different computational budgets. Table~\ref{tab:crowd_city} presents comprehensive results on CrowdHuman and CityPersons, organized from baseline through traditional and recent lightweight to higher-capacity models. This layout shows how our method compares within the ultra-lightweight regime and how far it closes the gap to models spending several times its budget. We focus on Recall since under heavy crowding missed pedestrians dominate error and most directly reflect robustness to occlusion.

\subsubsection{Comparison with Lightweight Detectors.}
Baseline YOLOv11n \cite{yolo11} achieves 79.2\% AP50 with 6.3 GFLOPs. Recent lightweight detectors show strong performance: YOLOv13n \cite{wang2025yolov13} reaches 79.2\% AP50 (6.4 GFLOPs); YOLOv12n \cite{zhang2025yolov12} achieves 79.4\% AP50 (6.5 GFLOPs); DETR-based D-FINE-N \cite{sun2025dfine} reaches 79.8\% AP50 while DEIM-D-FINE-N \cite{liu2025deim} achieves 80.1\% AP50, though requiring 7.0 GFLOPs and 4.0M parameters.

DensePed-Lite achieves 80.9\% AP50 and 50.3\% AP on CrowdHuman, outperforming DEIM-D-FINE-N by 0.8\% and 0.5\% while using only 5.5 GFLOPs and 2.53M parameters---roughly 60\% of its parameter count. Most telling is Recall: 71.3\%, a 3.5-point gain over baseline (67.8\%) and 2.5 points above DEIM-D-FINE-N, meaning we recover pedestrians competing methods miss entirely. Crucially, Precision simultaneously improves to 85.8\%. Both metrics rising together is the signature our design predicts: spatial completion and coherent fusion surface more occluded candidates while uncertainty-aware scoring prevents unreliable ones from inflating false positives. A method simply lowering confidence threshold would trade one for the other; obtaining both indicates gains stem from better feature quality rather than shifted operating point.

On CityPersons, the pattern holds: 62.1\% AP50 and 38.6\% AP, with Recall improving from 51.3\% to 52.9\% and Precision from 76.0\% to 79.7\%. That improvements transfer across two datasets with very different crowd densities suggests the underlying mechanism addresses occlusion-induced degradation generally rather than overfitting to one dataset's statistics.

\subsubsection{Efficiency-Accuracy Trade-off.}
Among high-capacity models, RT-DETR \cite{zhao2024rtdetr} achieves 80.8\% AP50 but requires 56.9 GFLOPs and 19.9M parameters (10.3$\times$ and 7.9$\times$ ours). YOLOv11s \cite{yolo11} reaches 81.4\% AP50 with 21.3 GFLOPs and 9.4M parameters.

\begin{figure}[t]
\centering\includegraphics[width=0.95\textwidth]{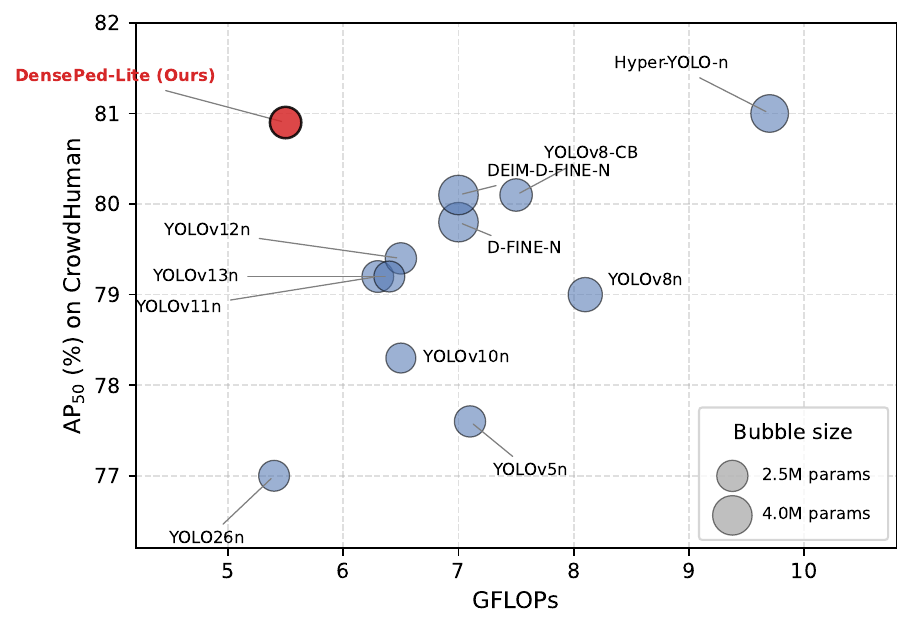}
\caption{Efficiency-accuracy trade-off for ultra-lightweight detectors on CrowdHuman. Bubble size represents parameter count. DensePed-Lite achieves the best AP50 with near-minimal GFLOPs.}
\label{fig:tradeoff}
\end{figure}

Figure~\ref{fig:tradeoff} visualizes the efficiency-accuracy landscape for ultra-lightweight detectors. DensePed-Lite occupies the Pareto-optimal frontier: highest AP50 (80.9\%) with second-lowest GFLOPs (5.5), marginally above YOLO26n's 5.4 but with 3.9 points higher accuracy. Among methods under 4M parameters, no competing detector matches our accuracy and efficiency simultaneously.

Mamba-YOLO-T reaches 81.8\% AP50 but spends 12.4 GFLOPs and 5.66M parameters---more than twice our budget---for a 0.9-point edge, and trails us on CityPersons Recall (50.8\% vs 52.9\%). Traditional detectors sit further out: Faster R-CNN and RetinaNet consume 251.4 and 199.7 GFLOPs yet do not surpass us on either AP50, confirming raw capacity is not what dense scenes reward. DensePed-Lite obtains best accuracy among all lightweight methods while spending the least, remaining competitive with models several times larger.

\subsection{Ablation Studies}

\subsubsection{Module Combination Analysis.}

To verify the compatibility and synergistic effects of MPSC (M), CTDM (T), and UQE (U), we conduct ablation studies by progressively adding these modules to the baseline framework.

Table~\ref{tab:module_ablation} shows that each module contributes on its own, but the more informative signal is how the variants rank after combination. Individually, UQE provides the largest gain (+1.7\% AP50) through better classification--localization alignment while \emph{reducing} FLOPs; MPSC adds +0.6\% AP50 by widening spatial coverage; and CTDM adds +0.9\% AP50 with a Precision gain from coherent fusion. The revealing observation is what happens in between: every two-module variant (+M+U, +T+U, +M+T) falls in a narrow 61.2--61.5\% AP50 band, barely separable from UQE alone (61.3\%), whereas only the complete model reaches 62.1\%. Thus, no pair is sufficient and the decisive jump appears only when all three breakpoints are closed. This supports our premise that the failure modes form a coupled cycle: leaving any one breakpoint open caps the benefit of the other two. The complete configuration also has the most efficient operating point (5.5 GFLOPs / 2.53M), indicating shared rather than duplicated computation.

\begin{table}[t]
\caption{Ablation Study Results (CityPersons). M: MPSC, U: UQE, T: CTDM.}
\label{tab:module_ablation}
\centering
\resizebox{\textwidth}{!}{%
\begin{tabular}{lccccc}
\hline
\textbf{Model} & \textbf{Recall} & \textbf{AP50(\%)} & \textbf{AP(\%)} & \textbf{Precision} & \textbf{GFLOPs / Params} \\
\hline
baseline & 0.513 & 59.6 & 36.1 & 0.760 & 6.3 / 2.58M \\
+ M        & 0.515 & 60.2 & 36.7 & 0.759 & 6.2 / 2.65M \\
+ U        & 0.525 & 61.3 & 37.4 & 0.776 & 5.6 / 2.42M \\
+ T        & 0.510 & 60.5 & 36.7 & 0.799 & 6.3 / 2.62M \\
+ M + U    & 0.519 & 61.2 & 37.6 & 0.774 & 5.5 / 2.50M \\
+ T + U    & 0.518 & 61.5 & 37.4 & 0.778 & 5.6 / 2.46M \\
+ M + T    & 0.519 & 61.3 & 36.9 & 0.777 & 6.2 / 2.69M \\
ours       & 0.529 & 62.1 & 38.6 & 0.797 & 5.5 / 2.53M \\
\hline
\end{tabular}%
}
\end{table}

The complete model achieves 62.1\% AP50 and 38.6\% AP, representing 2.5\% improvement over baseline while reducing GFLOPs by 12.7\%. The synergistic improvement exceeds individual module gains, validating the complementary design where MPSC provides spatial completeness, UQE ensures detection quality, and CTDM maintains multi-scale coherence.

\subsubsection{CTDM Module Analysis.}

CTDM comprises Token Statistics Self-Attention (T), DynamicTanh (D), and Mona normalization (M). We conducted ablation to understand their contributions.

The results in Table~\ref{tab:ctdm_ablation} reveal a pattern central to our design. Added in isolation, each component provides modest gains: +T, +D, and +M improve AP50 by 0.2\%, 0.1\%, and 0.3\% respectively, indicating individual components contribute but are insufficient alone. Pairing stabilization with verification (+T+D) achieves 60.2\% AP50, and the full stabilize-verify-refine loop reaches 60.5\% with a notable Precision gain to 79.9\%. This validates the closed-loop design: stabilization prepares distributions for reliable verification, verified global context guides local refinement, and refined features feed back to stabilization. Each stage is only effective in the presence of others, forming a coupled mechanism rather than independent enhancements.

\begin{table}[t]
\caption{Ablation Study Results of CTDM Module (CityPersons)}
\label{tab:ctdm_ablation}
\centering
\resizebox{\textwidth}{!}{%
\begin{tabular}{lccccc}
\hline
\textbf{Model} & \textbf{Recall} & \textbf{AP50(\%)} & \textbf{AP(\%)} & \textbf{Precision} & \textbf{GFLOPs / Params} \\
\hline
baseline   & 0.513 & 59.6 & 36.1 & 0.760 & 6.3 / 2.58M \\
+ T        & 0.515 & 59.8 & 36.3 & 0.768 & 6.3 / 2.56M \\
+ D        & 0.514 & 59.7 & 36.2 & 0.763 & 6.3 / 2.58M \\
+ M        & 0.516 & 59.9 & 36.4 & 0.771 & 6.4 / 2.64M \\
+ T + D    & 0.519 & 60.2 & 36.8 & 0.776 & 6.3 / 2.56M \\
Full CTDM  & 0.510 & 60.5 & 36.7 & 0.799 & 6.3 / 2.62M \\
\hline
\end{tabular}%
}
\end{table}

This dependency is exactly what the closed-loop design predicts. Verification alone, applied to skewed feature distributions, models global dependencies over unreliable statistics and amplifies rather than suppresses the noise from occluded regions; refinement alone sharpens boundaries without a trustworthy global context to anchor them, reinforcing whatever artifacts are already present; and stabilization alone merely rescales activations without using the stabilized signal for anything. Only when stabilization feeds verification, and verified context guides refinement, does each stage operate on inputs clean enough for it to help. The full loop lifts Precision by 3.9 points to 79.9\%, indicating sharper pedestrian-background separation in crowded scenes. This non-additive behavior is evidence that CTDM is a coupled mechanism rather than a stack of independent enhancements---which is precisely the property a reviewer should expect from a module designed to \emph{enforce} coherence rather than assume it.

\subsubsection{UQE Uncertainty Modeling Analysis.}

To validate that UQE genuinely models uncertainty rather than merely adding another feature channel, we compare three scoring strategies: baseline (classification confidence only), TopK pooling \cite{li2023generalizedfocal}, and our variance-based UQE. Table~\ref{tab:uqe_ablation} shows the results.

\begin{table}[t]
\caption{Ablation Study of UQE Uncertainty Modeling (CityPersons)}
\label{tab:uqe_ablation}
\centering
\resizebox{0.85\textwidth}{!}{%
\begin{tabular}{lcccc}
\hline
\textbf{Scoring Strategy} & \textbf{Recall} & \textbf{AP50(\%)} & \textbf{AP(\%)} & \textbf{Precision} \\
\hline
Baseline (cls only)       & 0.513 & 59.6 & 36.1 & 0.760 \\
+ TopK pooling           & 0.519 & 60.4 & 36.8 & 0.768 \\
+ Variance (no $\lambda$) & 0.521 & 60.8 & 37.1 & 0.771 \\
+ UQE (learnable $\lambda$) & 0.525 & 61.3 & 37.4 & 0.776 \\
\hline
\end{tabular}%
}
\end{table}

TopK pooling improves over baseline by 0.8\% AP50, capturing distribution sharpness but not spread. Adding variance without learnable weighting gains another 0.4\%, confirming that spread information helps. The full UQE with learnable $\lambda$ reaches 61.3\% AP50, demonstrating that the network learns to calibrate the uncertainty penalty to dataset-specific occlusion characteristics. The monotonic Precision improvement (76.0\% $\to$ 77.6\%) across variants confirms the penalty genuinely down-weights unreliable detections rather than simply shifting the confidence threshold.

\subsubsection{MPSC Path Redundancy Analysis.}

To verify that MPSC's gains stem from spatial path diversity rather than increased capacity, we compare: single-path baseline, dual-path without gating (equal weights), and full MPSC with adaptive gating. Table~\ref{tab:mpsc_ablation} reports the results.

\begin{table}[t]
\caption{Ablation Study of MPSC Path Configuration (CityPersons)}
\label{tab:mpsc_ablation}
\centering
\resizebox{0.85\textwidth}{!}{%
\begin{tabular}{lcccc}
\hline
\textbf{Configuration} & \textbf{Recall} & \textbf{AP50(\%)} & \textbf{AP(\%)} & \textbf{Precision} \\
\hline
Single-path baseline    & 0.513 & 59.6 & 36.1 & 0.760 \\
Dual-path (equal weight) & 0.514 & 59.9 & 36.4 & 0.758 \\
MPSC (adaptive gating)  & 0.515 & 60.2 & 36.7 & 0.759 \\
\hline
\end{tabular}%
}
\end{table}

Dual-path with equal weighting adds only 0.3\% AP50, indicating that naive redundancy provides limited benefit and even slightly hurts Precision (76.0\% $\to$ 75.8\%) due to noise amplification. Adaptive gating recovers Precision and reaches 60.2\% AP50, confirming that the network learns to down-weight corrupted paths under occlusion. This quality-aware path selection is what distinguishes MPSC from simply widening the network.

\subsection{Visualization Analysis}

\begin{figure}[t]
\centering
\subfigure[CrowdHuman dataset]{
    \includegraphics[width=0.95\textwidth]{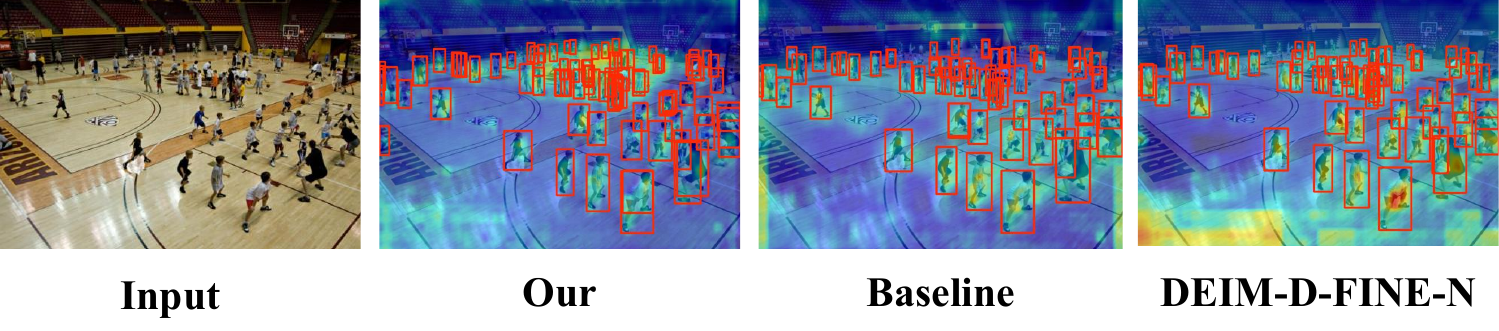}
    \label{fig:heatmap_crowd}
}
\subfigure[CityPersons dataset]{
    \includegraphics[width=0.95\textwidth]{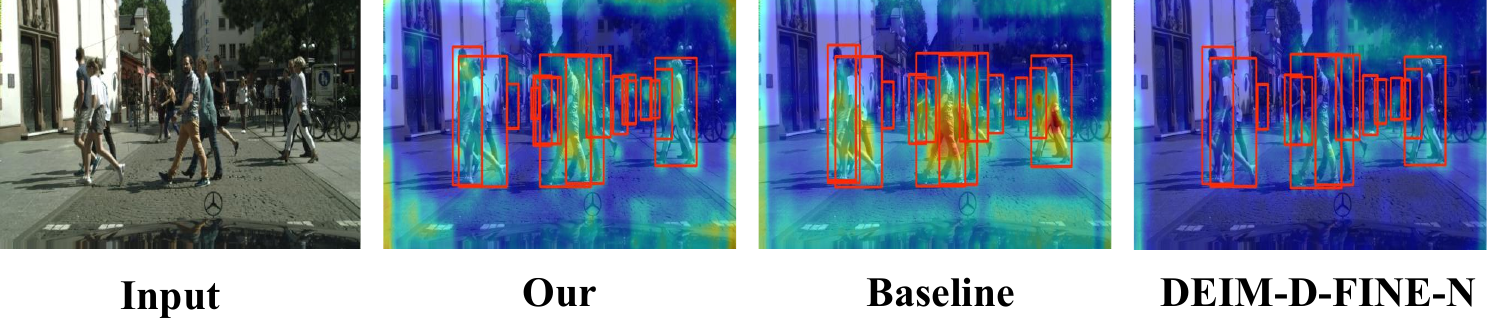}
    \label{fig:heatmap_city}
}
\caption{Grad-CAM visualization comparing baseline YOLOv11n, DEIM-D-FINE-N, and DensePed-Lite on heavily occluded scenarios.}
\label{fig:heatmap}
\end{figure}

As shown in Figure~\ref{fig:heatmap}, DensePed-Lite produces noticeably fewer missed detections and more sharply localized attention than both the baseline and DEIM-D-FINE-N. Two patterns are worth highlighting. On CrowdHuman, where bodies overlap heavily, the baseline's activation tends to merge neighbouring pedestrians into a single diffuse blob, whereas our method keeps the response concentrated on each individual even when only a fraction of the body is visible---consistent with the spatial completion and coherence-enforcement mechanisms recovering structure that a single scanning path would lose. On CityPersons, the differences appear most on partially visible pedestrians at image boundaries and behind obstacles, where the baseline either fires weakly or not at all while DensePed-Lite retains a confident, well-localized response; this matches the role of uncertainty-aware scoring in preventing such low-visibility but genuine targets from being suppressed. The visual evidence therefore aligns with the quantitative Recall gains: the improvements concentrate precisely on the occluded cases that motivated the design, rather than being spread uniformly across easy and hard instances.

\section{Conclusion}

We have presented DensePed-Lite, a lightweight detector built on a single principle: under occlusion a network should adapt its behaviour to the quality of what it observes rather than assume the input is complete. We traced the failures of lightweight detectors in dense scenes to three coupled effects---unreliable confidence, fragmented spatial coverage, and incoherent multi-scale fusion---that reinforce one another, and we realized the quality-aware principle at exactly these three points through uncertainty-aware scoring, multi-path spatial completion, and a stabilize--verify--refine fusion loop. Experiments on CrowdHuman and CityPersons show that this coordinated design attains the best accuracy among ultra-lightweight detectors while using only 5.5 GFLOPs and 2.53M parameters, with the gains concentrated on the high-Recall, occlusion-heavy regime the method targets. The ablations support the central claim: the three mechanisms are most effective together, and removing any one leaves a residual failure mode that limits the rest.

The study has limitations that point to future work. Our evaluation covers two pedestrian benchmarks; broader validation on other crowded categories and on embedded hardware would further establish practicality. The quality-aware principle is also not specific to pedestrians, and extending it to general dense-object detection is a natural next step.

\subsubsection*{Disclosure of Interests}
The authors have no competing interests to declare that are relevant to the content of this article.

%
%
\bibliographystyle{splncs04}
\bibliography{refs}

\end{document}